\documentclass[aps,pre,groupedaddress]{revtex4}
\usepackage{amsfonts}
\usepackage{mathrsfs}
\usepackage{graphicx}
\usepackage{amsmath}
\usepackage{amssymb}
\usepackage{amsbsy}
\usepackage{bm}

\begin{document}

\title{Relationship between manifold smoothness and adversarial vulnerability in deep learning with local errors}
\author{Zijian Jiang}
\author{Jianwen Zhou}
\author{Haiping Huang}
\affiliation{PMI Lab, School of Physics,
Sun Yat-sen University, Guangzhou 510275, People's Republic of China}
\date{\today}

\begin{abstract}
Artificial neural networks can achieve impressive performances, and even outperform humans in some specific tasks. Nevertheless, unlike biological brains, 
the artificial neural networks suffer from tiny perturbations in sensory input, under various kinds of adversarial attacks. It is therefore necessary to study the origin of 
the adversarial vulnerability. Here, we establish a fundamental relationship between geometry of hidden representations (manifold perspective) and 
the generalization capability of the deep networks. For this purpose, we choose a deep neural network trained by local errors, and then analyze emergent properties of trained networks
through the manifold dimensionality, manifold smoothness, and the generalization capability. To explore effects of adversarial examples, we consider independent Gaussian noise attacks and 
fast-gradient-sign-method (FGSM) attacks. Our study reveals that a high generalization accuracy requires a relatively fast power-law decay of the eigen-spectrum of hidden representations. Under Gaussian attacks,
the relationship between generalization accuracy and power-law exponent is monotonic, while a non-monotonic behavior is observed for FGSM attacks. Our empirical study provides a route towards a final 
mechanistic interpretation of adversarial vulnerability under adversarial attacks.
\end{abstract}

 \maketitle


\section{Introduction}
\label{Intro}
 Artificial deep neural networks have achieved the state-of-the-art performances in many domains such as pattern recognition and even natural language
 processing~\cite{DL-2016}. However, deep neural networks suffer from adversarial attacks~\cite{Carl-2017,Su-2019}, i.e., they can make an incorrect classification with high confidence when the input image
 is slightly modified yet maintaining its class label. In contrast, for humans and other animals, the decision making systems in the brain are quite robust to imperceptible 
 pixel perturbations in the sensory inputs~\cite{Zhou-2019}. This immediately establishes a fundamental question: what is the origin of the adversarial vulnerability of artificial 
 neural networks? To address this question, we can first gain some insights from recent experimental observations of biological neural networks.
 
A recent investigation of recorded population activity in the visual cortex of awake mice revealed a power law behavior
in the principal component spectrum of the population responses~\cite{Nature-2019}, i.e., the $n^{{\rm th}}$ biggest principal component (PC) variance 
scales as
$n^{-\alpha}$, where $\alpha$ is the exponent of the power law. In this analysis, the exponent is always slightly greater than one
for all input natural-image stimuli, 
reflecting an intrinsic property of a smooth coding in biological neural networks. It can be proved that when the exponent
is smaller than $1+2/d$, where $d$ is the manifold dimension of the stimuli set, 
the neural coding manifold must be fractal~\cite{Nature-2019}, and thus slightly modified inputs may cause extensive changes in outputs. In other words, 
the encoding in a slow decay of population variances would capture fine details of sensory inputs, rather than an abstract concept summarizing the inputs.
For a fast decay case, the population coding occurs in a smooth and differentiable manifold, and the dominant variance in the eigen-spectrum captures key features of the object
identity. Thus, the coding is robust, even under adversarial attacks. Inspired by this recent study, we ask whether the power-law behavior exists in
the eigen-spectrum of the correlated hidden neural activity
in deep neural networks. Our goal is to clarify the possible fundamental relationship between classification accuracy,
the decay rate of activity variances, manifold dimensionality and adversarial attacks of different nature.
 
Taking the trade-off between biological reality and theoretical analysis, we consider a special type of deep neural network,
trained with a local cost function at each layer~\cite{Mosta-2018}. Moreover, this kind of
training offers us the opportunity to look at the aforementioned fundamental relationship at each layer.
The input signal is transferred by trainable feedforward weights, while the error is propagated back to 
adjust the feedforward weights via random quenched weights connecting the classifier at each layer.
The learning is therefore guided by the target at each layer, and layered representations 
are created due to this hierarchical learning. These layered representations offer us the neural activity space for 
the study of the above fundamental relationship.

We remark that the motivation and relevance of our model setting, i.e., deep supervised learning with local errors.
As already known, the standard backpropagation widely used in machine learning is not biologically plausible~\cite{Back-2020}. The algorithm has three unrealistic (in biological sense)
assumptions: (i) errors are generated from the top layer and are thus non-local; (ii) a typical network is deep, thereby requiring a memory buffer for all layers' activities;
(iii) weight symmetry is assumed for both forward and backward passes. In our model setting, the errors are provided by local classifier modules and are thus local.
Updating the forward weight needs only the neural state variable in the corresponding layer [see Eq.~(\ref{leq})], without requiring the whole memory buffer.
And finally, the error is backpropagated through a fixed random projection, allowing easy implementation of breaking the weight symmetry. The learning algorithm in our paper thus bypasses 
the above three biological implausibilities~\cite{Mosta-2018}. Moreover, this model setting still allows deep network to transform the low-level features at earlier layers
into high-level abstract features at deeper layers~\cite{Yamin-2016,Mosta-2018}. Taken together, the model setting offers us the opportunity to look at
the fundamental relationship between classification accuracy, the power-law decay rate of activity variances, manifold dimensionality, and adversarial vulnerability \textit{at each layer}.
\begin{figure}
	\centering
	\includegraphics[bb=1 4 295 150,scale=1.1]{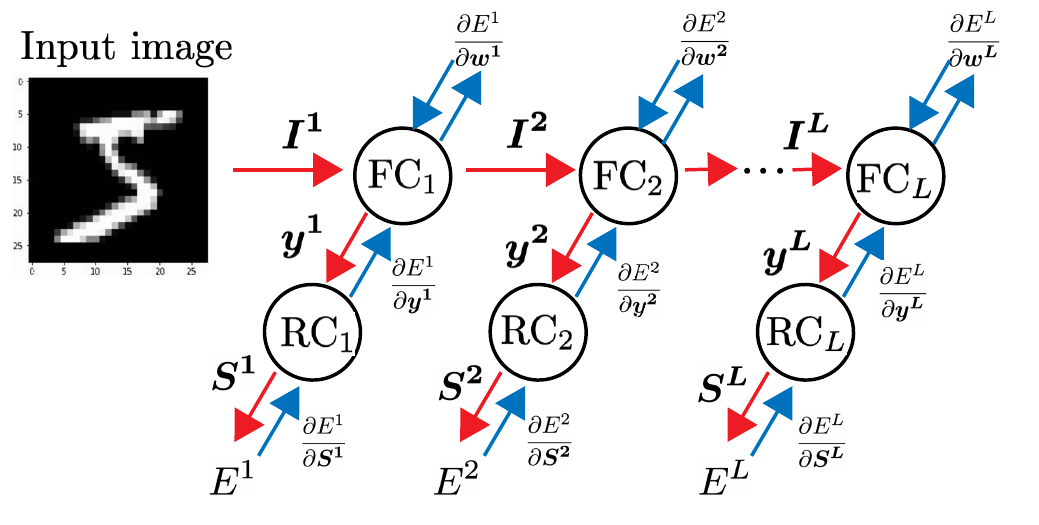}
	\caption{Schematic illustration of deep neural networks trained with local errors.
	FC is the short-hand for the fully-connected module, 
	which performs a non-linear processing. RC refers to the random classifier module, which 
	projects the output of FC to a 10-dimensional decision space. The red and blue arrows indicate
	forward and backward propagations, respectively. For the forward propagation, the input image is transfered through the FCs
	in a layer wise manner. The FC at each layer $l$ produces an activation $\boldsymbol{y}^l$, which serves as an input ($\boldsymbol{I}^{l+1}$) to the next layer, or is further
	weighted by quenched random weights of a RC ($\mathbf{S}^l$).
	After a random projection, the cross-entropy cost ($E^l$) is evaluated. The classification error is back-propagated to adjust the 
	trainable network parameters, through gradients of the cost function with respect to feedforward weights ($\boldsymbol{w}^l$) between consecutive
	layers.
	}
	\label{DNN}
\end{figure}
\section{Model}
\label{Model}
The deep neural network under investigation is shown in Fig.~\ref{DNN}. We train the network of $L+1$ layers (numbered from $0$) to classify the MNIST handwritten digits~\cite{Lecun-1998}. 
The first layer serves as an input layer, receiving a 784-dimensional vector from the MINST dataset. These sensory inputs are then transferred to the next layer in a feedforward manner.
The $l^{{\rm th}}$ ($l\ge1$) layer has $n_l$ neurons connected to a random classifier of a fixed size $10$, 
corresponding to the number of digit classes. The random classifier means that the connection weight to the classifier is pre-sampled from a zero-mean Gaussian distribution~\cite{Mosta-2018}.
The depth $L$ and each layer's width $n_l$ are flexible according to our settings. 
The strength of the connection or the weight between neuron $i$ 
at the layer $l-1$ and neuron $j$ at the upper layer $l$ is denoted by $w^l_{ij}$, which is trainable. Similarly, at the layer $l$, 
the weight between neuron $i$ at this layer and neuron $j$ at the neighboring random classifier is specified by $J^l_{ij}$, which is quenched. 

For the forward propagation, e.g., at the layer $l$, we have an input $\mathbf{I}^l$ and the pre-activation
is given by $z^l_i = \sum_j w^l_{ji}I^l_{j}$, simply summarizing the input pixels according to the corresponding weights. 
To ensure that the pre-activation 
is of the order one, a scaling factor is often applied to the pre-activation. However, the scaling is not necessary in our task. 
A non-linear transfer function is then applied to the pre-activation to obtain an activation defined by $y^l_i = f(z^l_i)$. 
Here, we use the rectified linear 
unit (ReLU) function as the non-linear function, i.e., $f(x)=\mbox{max}(0,x)$. The input of the next 
layer $l+1$ is the activation of layer $l$, except for the first layer where the input is a $784$-dimensional vector characterizing one handwritten digit. 
Meanwhile, the activation of the current layer is also fed to the random classifier, resulting in a classification score 
vector whose components $S^l_i = \sum_j J^l_{ji}y^l_{j}$. The score vector can be transformed into a classification probability by applying the softmax 
function, i.e., $P^{l}_{i}=e^{S^l_i}/\sum_j e^{S^l_j}$. $P^{l}_{i}$ can be understood as the probability of the input-label prediction.
To specify the cost function,
we first define $\mathbf{h}$ as the one-hot representation for the label of an image. More precisely, $h_i = \delta_{i,q}$ (Kronecker delta function), 
where $q$ 
is the digit number of the input image. Finally, the local error function at the layer $l$ is given by $E^l=-\sum_i h_i\ln P_i^l$. 
The local error is nothing but the cross entropy between $\mathbf{P}^l$ and $\mathbf{h}$. The forward propagation process can be
summarized as follows:
\begin{equation}
\begin{aligned}
&I_i^l = 
\begin{cases}
&y^{l-1}_i,l>1\\
&\text{the $i$-th pixel in an input},l=1
\end{cases}
\\
&z^l_i = \sum_j w_{ji}^l I_j^l,\\
&y^l_i = \mbox{max}(0,z^l_i),\\
&S_i^l = \sum_j J_{ji}^l y^l_j,\\
&P^l_i = \frac{e^{S^l_i}}{\sum_j e^{S^l_j}},\\
&E^l=-\sum_i h_i\ln P_i^l,
\end{aligned}
\end{equation}
where $h_i = \delta_{i,q}$ (Kronecker delta function) and $q$ is the digit label of the input image.

The local cost function $E^l$ is minimized when $h_i = P_i$ for every $i$. 
The minimization is achieved by the gradient decent method. 
The gradient of the local error with respect to the weight of the feedforward layer can be calculated by applying the 
chain rule, given by:
\begin{equation}
\label{leq}
\frac{\partial E^l}{\partial w_{ij}^l} =\left(\sum_k(P_k^l - h_k)J^l_{jk}\right )f'(z^l_j)I^l_i.
\end{equation}
Then, all trainable weights are updated by following $w^l_{ij}(t+1) = w^l_{ij}(t) - \eta{\frac{\partial E^l}{\partial w_{ij}^l}}$, where $t$ and $\eta$ indicate the learning step and the learning rate, respectively.
In our settings, $J^l_{ij}$ is a pre-fixed random Gaussian variables~\cite{Mosta-2018}. In practice, we use the mini-batch-based stochastic gradient descent method. The entire training dataset is divided into hundreds of mini-batches;
after sweeping through all mini-batches, an epoch of learning is completed.
The learning is then tested in an unseen/test dataset. For the standard MNIST dataset, the training set has $60\ 000$ images, and the test set 
has $10\ 000$ images. 
We can conclude whether a test image is correctly classified at the layer $l$ 
by comparing the position of the maximum component of the classifier 
score vector $\mathbf{S}^l$ with that of the one-hot label $\mathbf{h}$, i.e.,
the image is correctly classified at the layer $l$ if $\mbox{argmax}_{i} S_{i}^l = \mbox{argmax}_{i} h_{i}$. 
The test accuracy  ${\rm A}^l_t$ (t means test) at layer $l$ is then estimated by the number of correctly-classified images divided by the total number of images in the test set. 

After learning, the input ensemble can be transfered throughout the network in a layer-wise manner. Then, at each layer, the activity statistics can be analyzed by the eigen-spectrum of the correlation matrix (or covariance matrix).
We use principle component analysis (PCA) to obtain the eigen-spectrum, 
which gives variances along orthogonal directions in the descending order. For each input image, 
the population output of $n_l$ neurons at the layer $l$ can be thought of as a point in the $n_l$-dimensional activation space. 
It then follows that, for $k$ input images, the outputs can be seen as a cloud of $k$ points. The 
PCA first finds the direction with a maximal 
variance of the cloud, then chooses the second direction orthogonal to the first one, and so on. 
Finally, the PCA identifies $n_l$ orthogonal directions and $n_l$ corresponding variances. In our current setting, the $n_l$ eigenvalues of the the 
covariance matrix of the neural manifold explain $n_l$ variances. Arranging the $n_l$ eigenvalues
in the descending order leads to the eigen-spectrum whose behavior will be later analyzed in the next section.

To consider effects of adversarial examples, 
we add perturbations $\delta\mathbf{I}^l$ to the original input $\mathbf{I}^l$ at each layer. The perturbation is added in a layer-wise manner because of our model setting 
where a local classifier is present at each layer.
The original input is obtained by the layer-wise propagation of an image from the test set. We consider two kinds of additive perturbations: 
one is the Gaussian white noise and the other is the fast gradient sign method (FGSM) noise~\cite{Szegedy-2014,Good-2015}, representing black-box attacks and white-box attacks, 
respectively. Each component of the white noise is an \textit{i.i.d} random number drawn from zero mean Gaussian distribution with different variances (attack/noise strength),
and each component of 
the FGSM noise is taken from the gradient of the local cost function at each layer with
respect to the immediate input of this layer. These types of perturbations are given as follows:
\begin{subequations}
\begin{align}
\delta\mathbf{I}_{G}^l(\epsilon)& = \epsilon\cdot\mathbf{z},\\
\delta\mathbf{I}_{F}^l(\epsilon) &= \epsilon\cdot\mbox{sgn}\left(\frac{\partial E^l}{\partial \mathbf{I}^l}\right),
\end{align}
\end{subequations}
where $z_i\sim\mathcal{N}(0,1)$, $\epsilon$ denotes the perturbation magnitude, and $\mbox{sgn}(x)\equiv \frac{|x|}{x}$. In fact,
the FGSM attack can be thought of as an $\ell_\infty$ norm ball attack around the original input image.

\begin{figure}
	\centering
	\includegraphics[bb=5 2 416 283,scale=0.6]{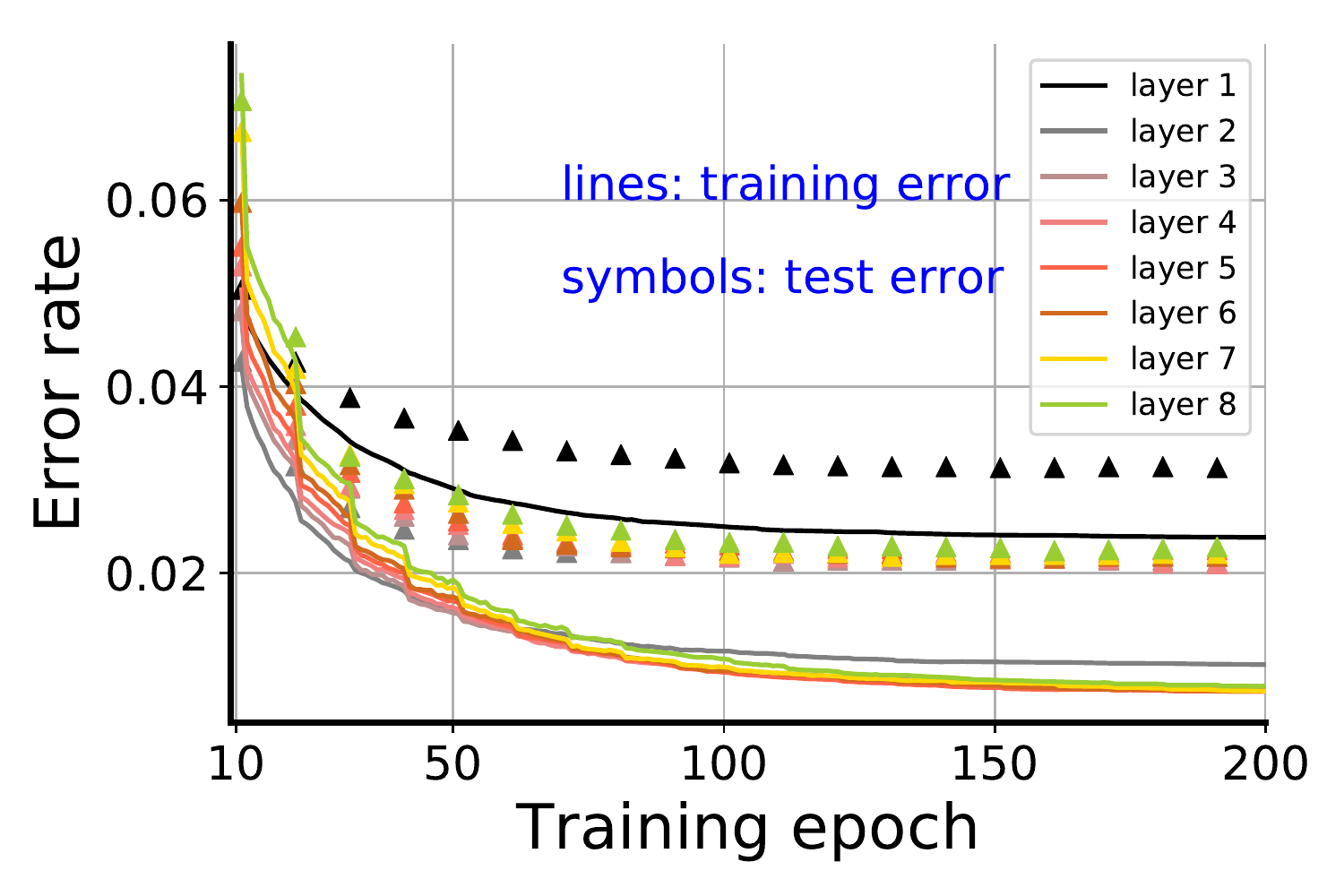}
	\caption{Typical trajectories of training and test error rates versus training epoch. 
	Lines indicate the train error rate,  and the symbols are the test error rate. The network width of each layer is fixed to $N=200$ (except the input layer),
	with $60\ 000$ images for training and $10\ 000$ images for testing. The initial learning rate $\eta=0.5$ which is multiplied by $0.8$ every ten epochs.
	}
	\label{training}
\end{figure}

\section{Results and Discussion}
\label{Res}
In this section, we apply our model to clarify the possible
fundamental relationship between classification accuracy, the decay rate of activity variances, manifold dimensionality
and adversarial attacks of different nature.
\subsection{Test error decreases with depth}
We first show that the deep supervised learning in our current setting works. Fig.~\ref{training} shows that the training error decreases as the test accuracy increases (before early stopping) during training.
We remark that it is challenging to rigorously prove the convergence of the algorithm we used in this study, as the deep learning cost landscape is highly non-convex, and 
the learning dynamics is non-linear in nature. As
a heuristic way, we judge the convergence by the stable error rate (in the global sense), which is also common in other deep learning systems.
As the layer goes deeper, the test accuracy grows until saturation despite a slight deterioration. This behavior provides an ideal candidate of deep learning to investigate the emergent properties of
the layered intermediate representations after learning, without and with adversarial attacks. Next, we will study in detail how the test accuracy is related to the power-law exponent, how the test accuracy is 
related to the attack strength, and how the dimensionality of the layered representation changes with the exponent, under zero, weak, and strong adversarial attacks.

\begin{figure}
	\centering
	\includegraphics[bb=7 3 851 284,scale=0.6]{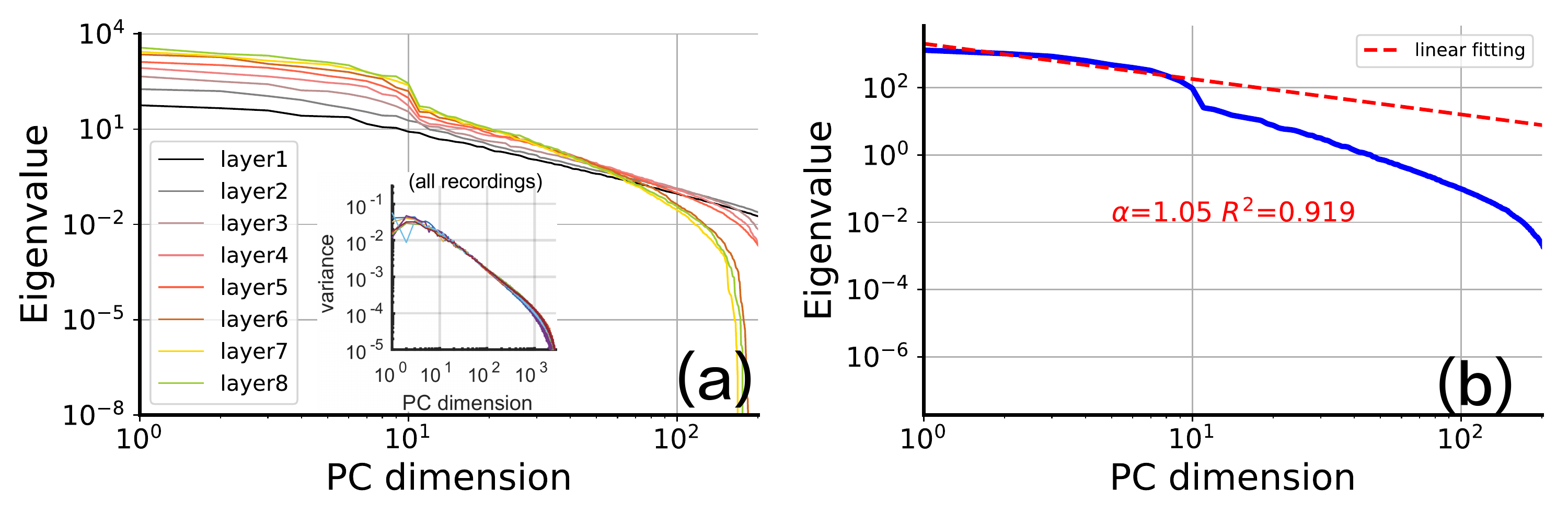}
	\caption{Eigen-spectrum of layer-dependent correlated activities and the power-law behavior of dominant PC dimensions.
	(a) The typical eigen-spectrum of deep networks trained with local errors ($L=8$, $N=200$). Log-log scales are used. 
	The inset is the eigen-spectrum measured in the visual cortex of mice (taken from Ref.~\cite{Nature-2019}). 
	(b) An example of extracting the power-law behavior at the fifth layer in (a). A linear fitting for the first ten PC components is shown in the log-log scale.
	}\label{fig_alpha}
\end{figure}

\subsection{Power-law decay of dominant eigenvalues of the activity correlation matrix}
A typical eigen-spectrum of our current deep learning model is given in Fig.~\ref{fig_alpha}. 
Notice that the eigen-spectrum is displayed in the log-log scale, then the slope of the linear fit of the spectrum gives
the power-law exponent $\alpha$. 
We use the first ten PC components to estimate $\alpha$ but not all for the following two reasons: 
($i$) A waterfall phenomenon appears at the position around the $10^{{\rm th}}$ dimension, which is more evident at higher layers. ($ii$)
The first ten dimensions explain more than $95\%$ of the total variance, and thus they capture the key information 
about the geometry of the representation manifold. The waterfall phenomenon in the eigen-spectrum can occur multiple times, especially for deeper layers [Fig.~\ref{fig_alpha} (a)], which is 
distinct from that observed in biological neural networks [see the inset of Fig.~\ref{fig_alpha} (a)]. This implies that the artificial deep networks may capture fine details of 
stimuli in a \textit{hierarchical} manner. A typical example of obtaining the power-law exponent is shown in Fig.~\ref{fig_alpha} (b) for the fifth layer.
When the stimulus size $k$ is chosen to be large enough (e.g., $k\ge2000$; $k=3000$ throughout the paper), the fluctuation of the estimated exponent due to stimulus selection can be neglected. 
\begin{figure}
	\centering
	\includegraphics[bb=5 4 850 283,scale=0.6]{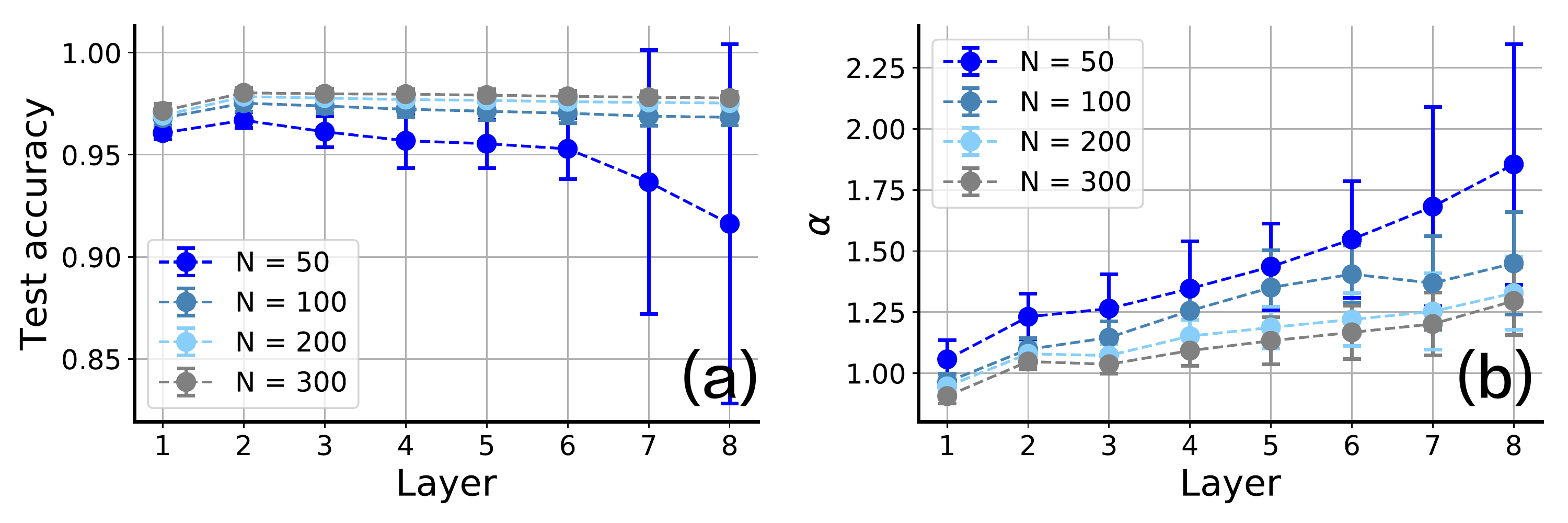}
	\caption{Effects of network width on test accuracy and power-law exponent $\alpha$. (a) Test accuracy versus layer. 
	Error bars are estimated over $20$ 
	independently training models. (b) $\alpha$ versus layer. Error bars 
	are also estimated over $20$ 
	independently training models.
	}
	\label{fig_acc}
\end{figure}
\subsection{Effects of layer width on test accuracy and power-law exponent}
We then explore the effects of layer width on both test accuracy and power-law exponent. As shown in Fig.~\ref{fig_acc} (a), the test accuracy becomes more stable with increasing layer width.
This is indicated by an example of $n_l=50$ which shows a large fluctuation of the test accuracy especially at deeper layers. We conclude that a few hundreds of neurons at each layer is sufficient for
an accurate learning.

The power-law exponent also shows a similar behavior; the estimated exponent shows less fluctuations as the layer width increases. This result also shows that the exponent grows with layers.
The deeper the layer is, the larger the exponent becomes. A larger exponent suggests that the manifold is smoother, because the dominant variance decays fast, leaving few space for encoding the irrelevant features in
the stimulus ensemble. This may highlight the depth in hierarchical learning is important for capturing key characteristics of sensory inputs.

\begin{figure}
	\centering
	\includegraphics[bb=7 10 421 283,scale=0.8]{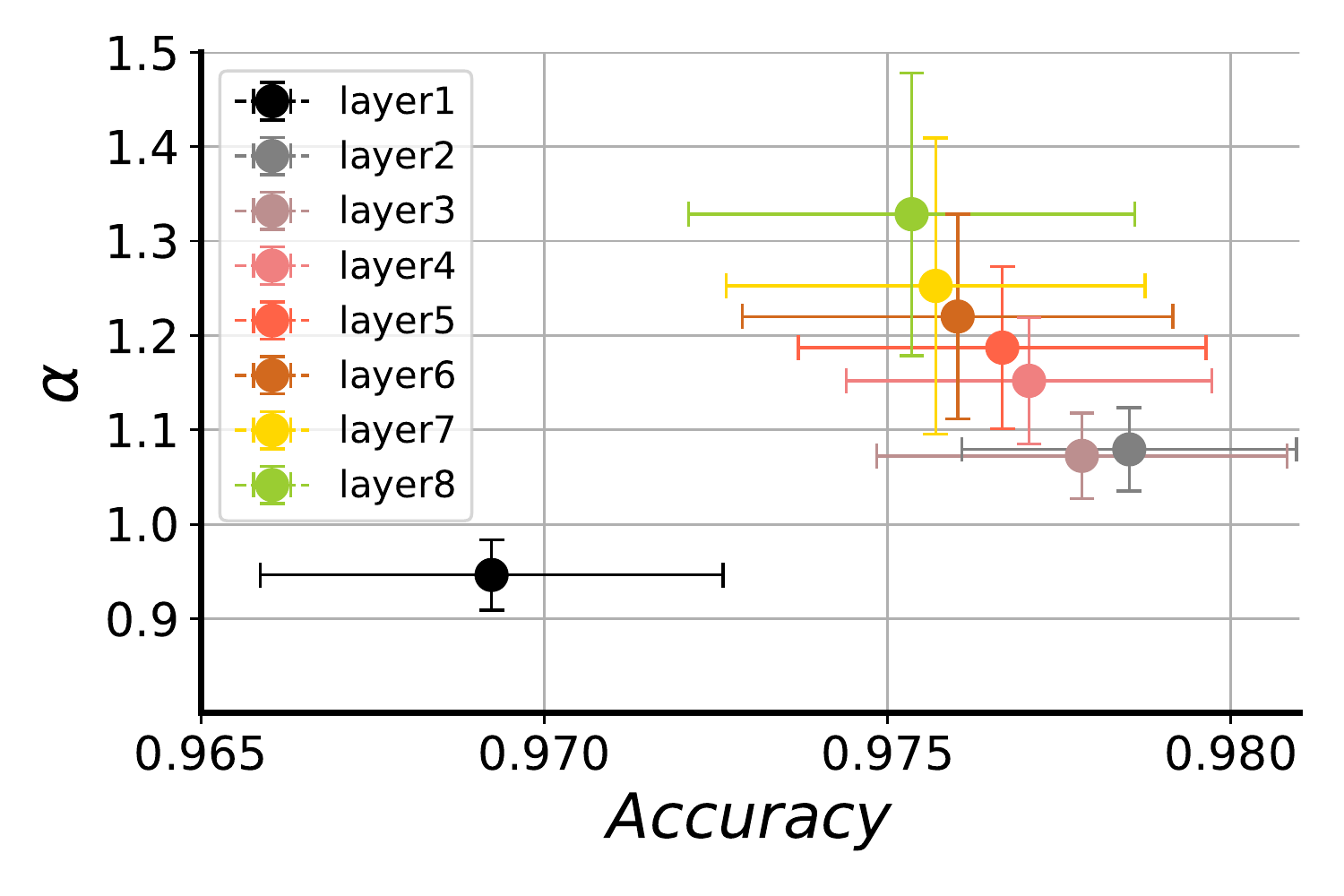}
	\caption{ The power-law exponent $\alpha$ versus test accuracy of the manifold. $\alpha$ grows along the depth, 
	while the test accuracy has a turnover at the layer 2, and then decreases by a very small margin. Error bars 
	are estimated over $50$ 
	independently training models.
	}
	\label{fig_alpha_acc}
\end{figure}
\subsection{Relationship between test accuracy and power-law exponent}
In the following simulations, 
we use $200$ neurons at each layer.
The relationship between
$\alpha$ and ${\rm A}_t^l$ is plotted in Fig.~\ref{fig_alpha_acc}, with 
error bars indicating the standard errors across $50$ independent trainings.
The typical value of $\alpha$ grows along the depth, 
implying that the coding manifold becomes smoother at higher layers. The clear separation of the $\alpha$ values between 
the earlier layer (e.g., the exponent of the first layer is less than one) and the deeper layers may indicate a transition from 
non-smooth (a slow decay of the eigen-spectrum) coding to smooth coding.
Interestingly, the test accuracy does not increase with the depth for deeper layers, just simply increasing first and then quickly
drops with a relatively small margin. On which layer the accuracy starts to be saturated seems to depend on the specific computational task.
This suggests that too much smoothness will harm the generalization ability of the network, 
although the impact is not significant and thus limited.

\begin{figure}
	\centering
	\includegraphics[bb=3 5 842 561,scale=0.5]{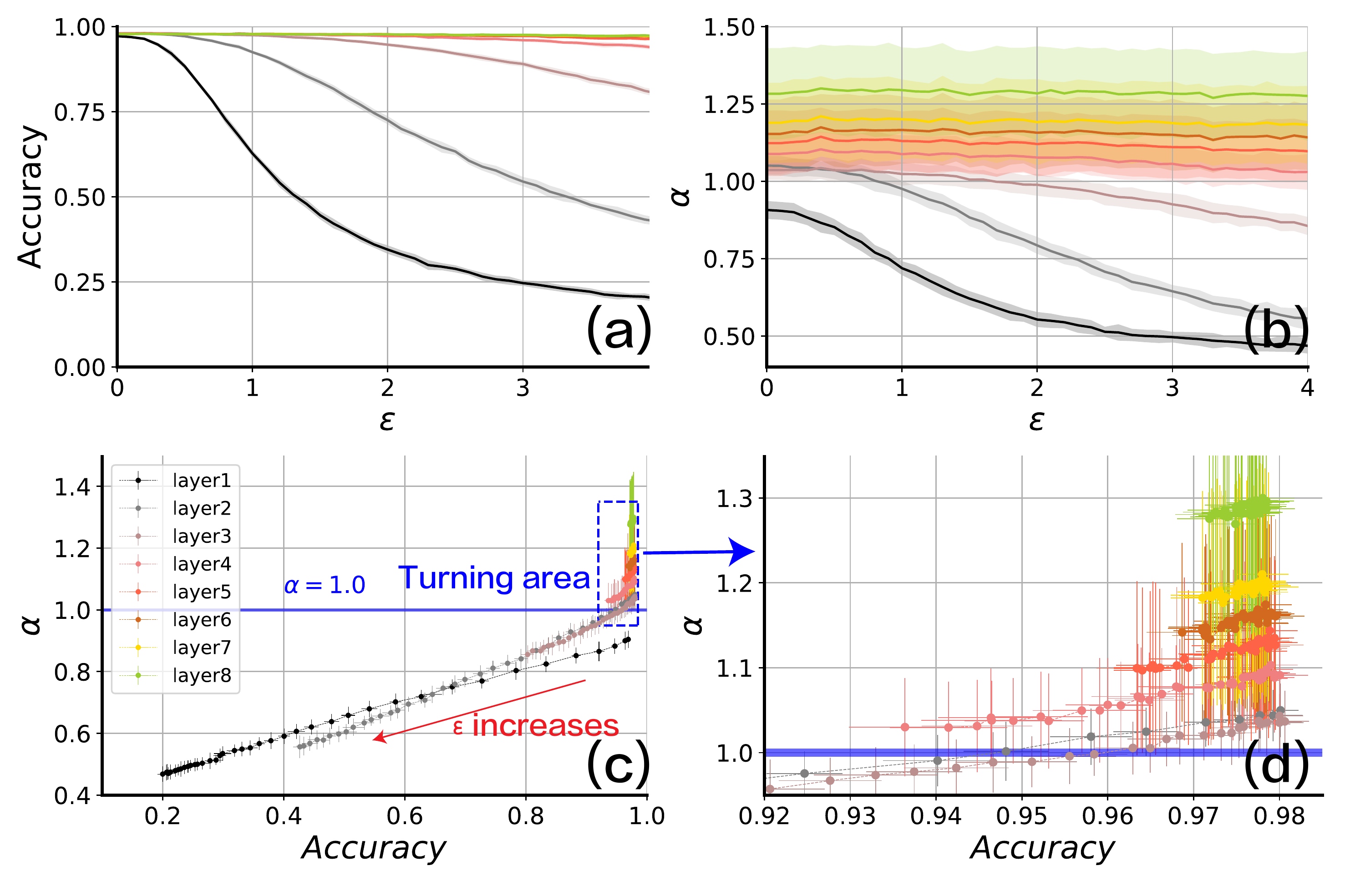}
	\caption{ Relationship between test accuracy and power-law exponent $\alpha$ when the input test data is attacked by
	independent Gaussian white noises. Error bars 
	are estimated over $20$ 
	independently training models. (a) Accuracy versus $\epsilon$. 
	$\epsilon$ is the attack amplitude. (b) $\alpha$ versus $\epsilon$. (c) Accuracy versus $\alpha$ over 
	different values of $\epsilon$. Different symbol colors refer to different layers. The red arrow points to the direction 
	along which $\epsilon$ increases from $0.1$ to $4.0$, with an increment size of $0.1$. 
	The relationship of ${\rm A}_t^l(\alpha)$ with increasing $\epsilon$ in the first 
	three layer show a linear function, with the slopes of $0.56$, $0.86$, and $1.04$ respectively.
	The linear fitting coefficients $R^2$ are all larger than $0.99$. Beyond the third layer, the linear relationship is not evident. 
	For the sake of visibility, we enlarge the deeper-layer region in (d). A turning 
	point $\alpha\approx1.0$ appears. Above this point, the manifold seems to become smooth, and the exponent becomes stable even against stronger
	black-box attacks [see also (b)].
	}
	\label{fig_GW}
\end{figure}
\subsection{Properties of the model under black-box attacks}
We first consider the additive Gaussian white noise perturbation to the input representation at each layer. This kind of perturbation is also called 
the black-box attack, because it is not necessary to have access to the training details of the deep learning model, including architectures and loss function.
Under this attack, the test accuracy decreases as the perturbation magnitude increases. As expected from the above results of manifold smoothness at deeper layers,
deeper layers become more robust against the black-box attacks of increasing perturbation magnitude [Fig.~\ref{fig_GW} (a, b)]. The $\epsilon$-dependence of the power-law exponent
shows that the adversarial robustness correlates with the manifold smoothness at deeper layers, which highlights the fundamental relationship between the adversarial vulnerability
and the fractal manifold~\cite{Nature-2019}.

Inspecting carefully the behavior of $\alpha$ as a function of the test accuracy, we can identify two interesting regimes separated by
a turning point at $\alpha\sim1$ [Fig.~\ref{fig_GW} (c)]. Below the turning point, increasing $\epsilon$ results in a clear decreasing
in both ${\rm A}_t^l$ and $\alpha$; Moreover, for the first three layers, a linear relationship can be identified. The linear fitting coefficient is as high as $R^2\sim0.99$.
This observation is very interesting. The underlying mechanism is closely related to the monotonic behavior of the accuracy or exponent as a function of attack strength.
Let us define their respective functions as $f_a(\epsilon)$ and $f_e(\epsilon)$. Then a simple transformation leads to $\alpha=f_e(f_a^{-1}({\rm A}_t^l))$, which suggests that a particular choice of 
$f_a$ and $f_e$ allows for the observed linear relationship, e.g., a linear function. Therefore, how the earlier layer is affected by adversarial examples with increasing strength plays a vital role
in shaping the interesting linear regime. Above the turning point, the linear fitting fails, and instead, a non-linear relationship takes over. One key reason is that
deeper layers are more robust against adversarial examples in our current setting.

\begin{figure}
	\centering
	\includegraphics[bb=3 5 860 860,width=0.6\textwidth]{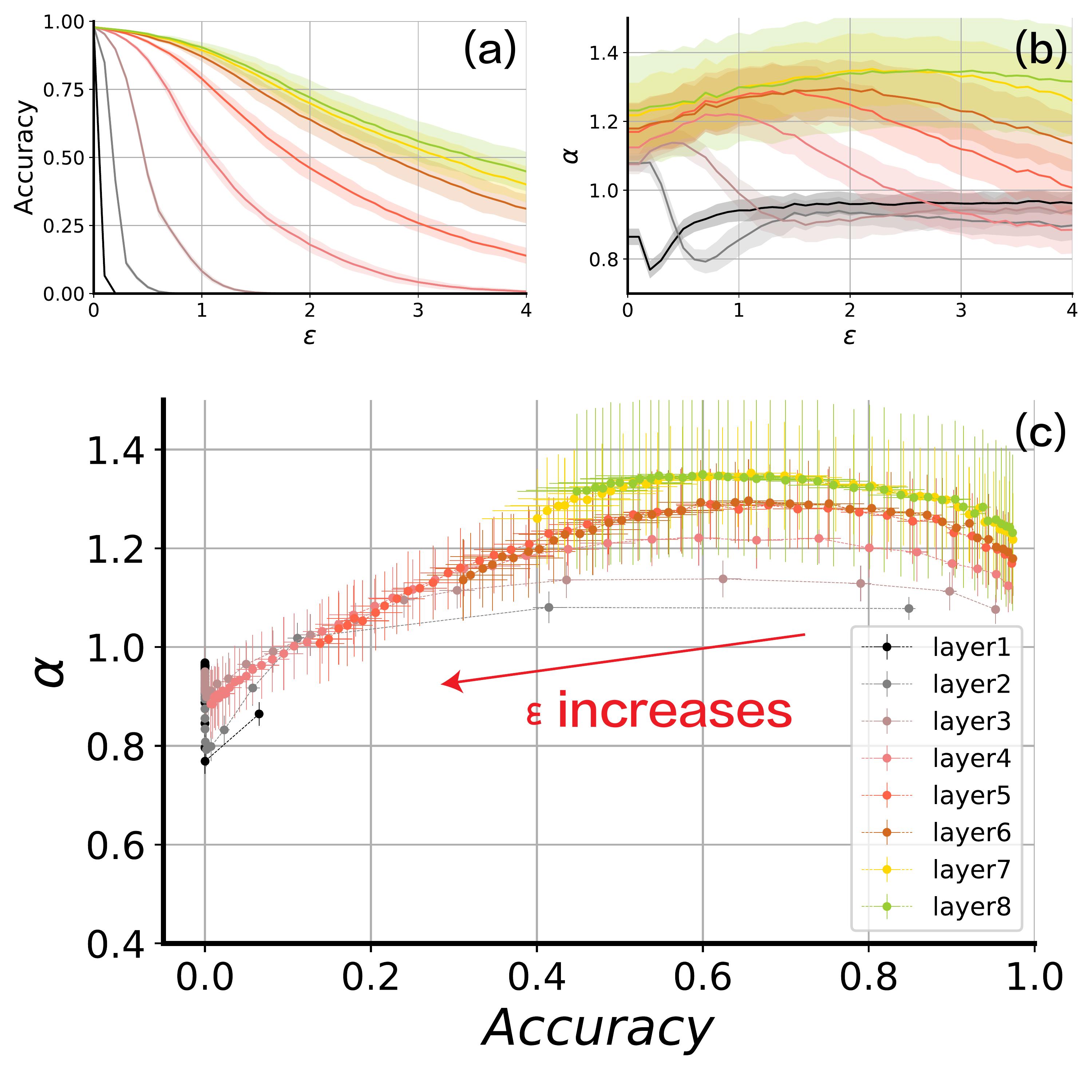}
	\caption{ Relationship between test accuracy and exponent $\alpha$ under the FGSM attack. Error bars 
	are estimated over $20$ 
	independently training models.
	(a) ${\rm A}_t^l$ changes with $\epsilon$. (b) $\alpha$ changes with $\epsilon$. 
	(c) ${\rm A}_t^l$ versus $\alpha$ over different attack magnitudes. $\epsilon$ increases from $0.1$ to $4.0$ 
	with the increment size of $0.1$. The plot shows a non-monotonic behavior different from that of white-box attacks in Fig.~\ref{fig_GW} (c).
	}
	\label{fig_FGSM}
\end{figure}

\begin{figure}
	\centering
	\includegraphics[bb=3 5 422 810,scale=0.57]{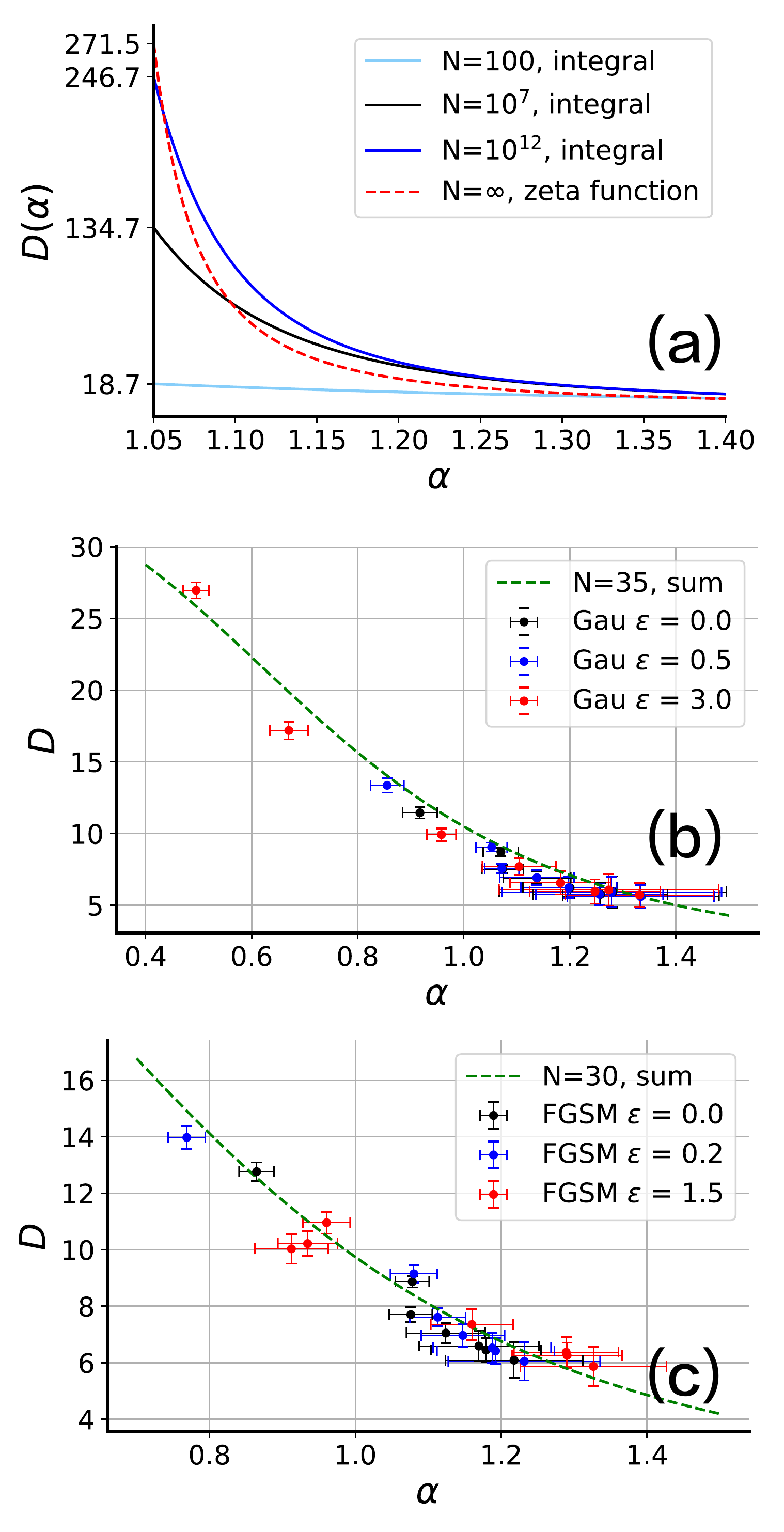}
	\caption{Relationship between dimensionality $D$ and power-law exponent. 
	(a) $D(\alpha)$ estimated from the integral approximation and in the thermodynamic limit.
	 $N$ is the layer width. (b) $D(\alpha)$ 
	under the Gaussian white noise attack. The dimensionality and the exponent are estimated directly from the layered 
	representations given the immediate perturbed input for each layer [Eq.~(\ref{dim1})]. We show three typical cases of attack: no noise with 
	$\epsilon=0.0$, small noise with $\epsilon=0.5$ and strong noise with $\epsilon=3.0$. For each 
	case, we plot eight results corresponding to eight layers. The green dashed line is the theoretical prediction [Eq.~(\ref{dim2})], provided that
	$N=35$. Error bars 
	are estimated over $20$ 
	independently training models.
	(c) $D(\alpha)$ under the FGSM attack. The theoretical curve (dashed line) is computed with $N=30$. Error bars 
	are estimated over $20$ 
	independently training models.
	}
	\label{fig_dim}
\end{figure}
\subsection{Properties of the model under white-box attacks}
We then consider the white-box attack---FGSM additive noise perturbation; the results are summarized in Fig~\ref{fig_FGSM}. The FGSM attack is much stronger than the black-box attack, as expected from
the fact that the loss function and network architecture knowledge are both used by the FGSM attack. The first few layers display evident adversarial vulnerability, or completely fail to identity correct class of the input images,
while the last deeper layers still 
show adversarial robustness to some extent. These deeper layers also maintain a relatively high value of $\alpha$, although strong adversarial examples strongly reduce the manifold smoothness especially for layers
next to the earlier layers. Under the FGSM attack, the $\alpha$-${\rm A}_t^l$ relationship becomes complicated, and the linear-nonlinear separation that occurs in the black-box attacks disappears.
However, the high accuracy still implies a high value of $\alpha$. In particular, when the white-box perturbation is weak, the system reduces the manifold smoothness by a small margin
to get a high accuracy [Fig.~\ref{fig_FGSM} (c)]. This can be interpreted as follows. Supported by the trained weights (no adversarial training), the manifold formed by the adversarial examples takes 
into account more details (or some special directions pointing to the decision boundary) of the adversarial examples. Under the FGSM attack, specific pixels with $\ell_\infty$ norm perturbations
affecting strongly the loss function are particularly used to flip the decision output. In this sense, the competition (or trade-off) between the accuracy and the manifold smoothness captured by
$\alpha$ is present. This may explain why there exists a peak in Fig.~\ref{fig_FGSM} (c). The peak also appears in Fig.~\ref{fig_FGSM} (b). Both types of peaks have the one-to-one correspondence. Note that black-box attacks has no such properties.

\subsection{Relationship between manifold linear dimensionality and power-law exponent}
The linear dimensionality of a manifold formed by data/representations can be thought of as a first approximation of intrinsic geometry of a manifold~\cite{Huang-2018,Huang-2020weak},
defined as follows:
\begin{equation}\label{dim1}
 D^l=\frac{\left(\sum_{i=1}^N\lambda_i\right)^2}{\sum_{i=1}^N\lambda_i^2},
\end{equation}
where $\{\lambda_i\}$ is the eigen-spectrum of the covariance matrix. Suppose the eigen-spectrum has a power-law decay behavior as the PC dimension increases,
we simplify the dimensionality equation as follows:
\begin{equation}\label{dim2}
 D(\alpha)=\frac{\left(\sum_{n=1}^Nn^{-\alpha}\right)^2}{\sum_{n=1}^Nn^{-2\alpha}}\simeq\frac{(N^{1-\alpha}-1)^2(1-2\alpha)}{(N^{1-2\alpha}-1)(1-\alpha)^2},
\end{equation}
where $N$ denotes the layer width, and
the approximation $\sum_{n=1}^Nn^{-\alpha}\simeq\int_1^Nn^{-\alpha}dn=\frac{N^{1-\alpha}-1}{1-\alpha}$ is used to get the second equality. 
In the thermodynamic limit, $D(\alpha)=\frac{[\varsigma(\alpha)]^2}{\varsigma(2\alpha)}$, where $\varsigma(\alpha)$ is the Reimann zeta function.
Note that for a small value of $N$, a theoretical prediction of $D(\alpha)$ in Fig.~\ref{fig_dim} (b,c) is obtained by using the first equality (sum).

Results are shown in Fig.~\ref{fig_dim}. The theoretical prediction agrees roughly with simulations under zero, weak and strong attacks of black-box and white-box types.
This shows that using the power-law decay behavior of the eigen-spectrum in terms of the first few dominant dimensions to study the relationship between the manifold geometry and 
adversarial vulnerability of artificial neural networks is also reasonable, as also confirmed by many aforementioned non-trivial properties about this fundamental relationship. Note that when the network width increases,
a deviation may be observed due to the waterfall phenomenon observed in the eigen-spectrum (see Fig.~\ref{fig_alpha}).

\section{Conclusion}
In this work, we study the fundamental relationship between the adversarial robustness and the manifold smoothness characterized by the power-law exponent of the eigen-spectrum. The eigen-spectrum is obtained from the correlated 
neural activity on the representation manifold. We choose deep supervised learning with local errors as our target deep learning model, because of its nice property of allowing for analyzing each layered representation in terms of both
test accuracy and manifold smoothness. We then reveal that the deeper layer has a larger value of $\alpha$, thereby possessing the adversarial robustness against both black- and white-box attacks.
In particular, we find a turning point under the black-box attacks for the ${\rm A}_t^l(\alpha)$ curve, separating linear and non-linear relationships.
This turning point also signals the qualitative change of the manifold geometric property. Under the white-box attacks, the exponent-accuracy behavior becomes more complicated, as isotropic properties of attacks like in 
Gaussian white noise perturbations do not hold, which requires that a trade-off between manifold smoothness and test accuracy, given the normal trained weights (no adversarial training), should be taken.

All in all, although our study does not provide precise mechanisms underlying the adversarial vulnerability, the empirical works are expected to offer some intuitive arguments 
about the fundamental relationship between generalization capability and the intrinsic properties of representation manifolds inside the deep neural networks with biological plausibility (to some degree), encouraging future mechanistic studies towards
the final goal 
of aligning machine perception and human perception~\cite{Zhou-2019}.

\begin{acknowledgments}
This research was supported by the National Key R$\&$D Program of China (2019YFA0706302), the National Natural Science Foundation of China for
Grant No. 11805284, and the start-up budget 74130-18831109 of the 100-talent-
program of Sun Yat-sen University.
\end{acknowledgments}


\end{document}